\documentclass{llncs}
\usepackage{times}
\usepackage{helvet}
\usepackage{courier}
\usepackage{footmisc}
\usepackage{makeidx}  
\usepackage{xcolor}
\usepackage{adjustbox}
\usepackage{amsmath}
\usepackage{amsfonts}
\usepackage{algorithmicx}
\usepackage{algorithm}
\usepackage{algpseudocode}

\usepackage{grffile}
\usepackage{hyperref}
\usepackage{xcolor}
\usepackage{adjustbox}
\usepackage{changepage}
\usepackage{subfig}
\usepackage[]{caption}
\usepackage{graphicx} 
\usepackage{amssymb}
\begin{document}

\makeatletter
\renewenvironment{thebibliography}[1]
     {\section*{\refname}
      \small
      \list{}%
           {\settowidth\labelwidth{}%
            \leftmargin\parindent
            \itemindent=-\parindent
            \labelsep=\z@
            \if@openbib
              \advance\leftmargin\bibindent
              \itemindent -\bibindent
              \listparindent \itemindent
              \parsep \z@
            \fi
            \usecounter{enumiv}%
            \let\p@enumiv\@empty
            \renewcommand\theenumiv{}}%
      \if@openbib
        \renewcommand\newblock{\par}%
      \else
        \renewcommand\newblock{\hskip .11em \@plus.33em \@minus.07em}%
      \fi
      \sloppy\clubpenalty4000\widowpenalty4000%
      \sfcode`\.=\@m}
     {\def\@noitemerr
       {\@latex@warning{Empty `thebibliography' environment}}%
      \endlist}
      \def\@cite#1{#1}%
      \def\@lbibitem[#1]#2{\item[]\if@filesw
        {\def\protect##1{\string ##1\space}\immediate
      \write\@auxout{\string\bibcite{#2}{#1}}}\fi\ignorespaces}
\makeatother

\newcommand\numberthis{\addtocounter{equation}{1}\tag{\theequation}}

\makeatletter
\newcommand\footnoteref[1]{\protected@xdef\@thefnmark{\ref{#1}}\@footnotemark}
\makeatother

\mainmatter              
\title{Crossprop: Learning Representations by Stochastic Meta-Gradient Descent in Neural Networks}
\titlerunning{Crossprop}  
%
\author{Vivek Veeriah\footnote{\label{note}These authors contributed equally to this work.}, Shangtong Zhang\footref{note}, Richard S. Sutton}
%
\authorrunning{Vivek Veeriah et al.} 
%
\tocauthor{Vivek Veeriah, Shangtong Zhang, Richard S. Sutton}
\institute{Dept. of Computing Science, \\ University of Alberta, Edmonton, AB, Canada\\
\email{\{ vivekveeriah, shangtong.zhang, rsutton \}@ualberta.ca}}

\maketitle              

\begin{abstract}
Representations are fundamental to artificial intelligence. The performance of a learning system depends on the type of representation used for representing the data. Typically, these representations are hand-engineered using domain knowledge. More recently, the trend is to {\em learn} these representations through stochastic gradient descent in multi-layer neural networks, which is called {\em backprop}. Learning the representations directly from the incoming data stream reduces the human labour involved in designing a learning system. More importantly, this allows in scaling of a learning system for difficult tasks. In this paper, we introduce a new incremental learning algorithm called {\em crossprop}, which learns incoming weights of hidden units based on the meta-gradient descent approach, that was previously introduced by Sutton (1992) and Schraudolph (1999) for learning step-sizes. The final update equation introduces an additional memory parameter for each of these weights and generalizes the backprop update equation. From our experiments, we show that crossprop learns and reuses its feature representation while tackling new and unseen tasks whereas backprop relearns a new feature representation.
\end{abstract}
\section{Introduction}

The type of representation used for presenting the data to a learning system plays a key role in artificial intelligence and machine learning. Typically, the performance of a learning system, such as its speed of learning or its error rate, directly depends on how the data is represented internally by the learning system. Hand-engineering these representations using some special domain knowledge was the norm for designing learning systems. More recently, these representations are learned hierarchically and directly from the data through stochastic gradient descent. Learning such representations significantly improves the performance of the learning system and reduces the human effort involved in designing a learning system. Importantly, this allows in scaling up of the learning systems for bigger and harder problems. 

Learning hierarchical representations directly from the data has recently gained a lot of popularity. Designing deep neural networks has allowed the learning systems to tackle incredibly hard problems: classifying or recognizing the objects from natural scene images (Deng et al., 2009; Szegedy et al., 2016), automatically translating text and speeches (Cho et al., 2014; Bahdanau et al., 2014; Wu et al., 2016), achieving and surpassing human-level baseline in Atari (Mnih et al., 2015), achieving super-human performance in Poker (Morav{\v c}{\' i}k et al., 2017) and in improving robot control from learning experiences (Levine et al., 2016). It is important to note that in many of these problems it is difficult to hand-engineer a data representation and an inadequate representation generally limits the performance or the scalability of the learning system.

The algorithm behind the training of such deep neural networks is called {\em backprop} (or {\em backpropagation}), which was introduced by Rumelhart, Hinton and Williams (1988). It extended the stochastic gradient descent approach, via chain rule, for learning the weights in the hidden layers of a neural network. 

Though backprop has produced many successful results, it suffers from some fundamental issues which makes it slow in learning a useful representation that solves many tasks. Specifically, backprop tends to interfere with the previously learned representations because the units that have so far been found to be useful are the ones that are most likely to be changed (Sutton, 1986). One of the reasons for this is that the weights of each hidden layer is assumed to be independent with each other, and because of this, the parameters of the neural network race against each other to minimize the error for a given example. In order to overcome this issue, the neural network needs to be trained over multiple sweeps (epochs) with the data so that algorithm can settle down with one representation that encompasses all the data it has seen so far.  

In this paper, we introduce a meta-gradient descent approach for learning the weights connecting the hidden units of a neural network. Previously, the meta-gradient descent approach was introduced by Sutton (1992) and Schraudolph (1999) for learning parameter-specific step-sizes, which is adapted here for learning the incoming weights that connect to the hidden units. Our proposed method is called {\em crossprop}.

This specifically addresses the racing problem which is observed in backprop. Furthermore, from our continual learning experiments where a learning system experiences a sequence of related tasks, we observed that crossprop tends to find the features that best generalize across these multiple tasks. Backprop, on the other hand, tends to {\em unlearn} and {\em relearn} the features with each task that it experiences. From a continual learning perspective, where a learning system experiences a sequence of tasks that are related with each other, it is desirable to have a learning system that can leverage its learning from its past experiences for solving unseen and more difficult tasks that it experiences in its future.

\section{Related Methods}

There are three fundamental approaches for learning representations, via a neural network, directly from the data. 

The first, and the most popular, approach for learning such representations is through stochastic gradient descent over the supervised learning error function, like the mean squared or the cross-entropy error (Rumelhart et al., 1988).  This approach is proved successful in many successful applications, ranging from difficult problems in computer vision to patient diagnoses. Although this method has a strong track record, it is not perfect yet. Particularly, learning representations by backpropagating the supervised error signal often learns slowly and poorly in many problems (Sutton, 1986; Jacobs, 1988). In order to address this, many modifications to backprop are introduced, like adding momentum (Jacobs, 1988), RMSProp (Tieleman and Hinton, 2012), ADADELTA (Zeiler, 2012), ADAM (Kingma and Ba, 2014) etc. and its not quite clear which variation of backprop will work well for a given task. However, all these variations of backprop still tend to interfere with the previously learned representations, thereby causing the network to unlearn and relearn representation even when the task can be solved by leveraging the learning from previous experiences. 

Another promising approach for learning representations is by the generate and test process (Klopf and Gose, 1969; Mahmood and Sutton, 2013). The underlying principle behind these approaches is to generate many features in a random manner and then test the usability for each of these features. Based on certain heuristics, the features are either preserved or discarded.  Furthermore, the generate and test approach can be combined with backprop to achieve a better rate of learning in supervised learning tasks. The primary motivation behind these generate and test approaches is to design a distributed and a computationally inexpensive representation learning method.

Some researchers have also looked at learning representations that fulfil certain unsupervised learning objectives, like clustering, sparsity, statistic independence or reproducibility of data, which takes us to the third fundamental approach towards learning representations (Olshausen and Field 1997; Comon, 1994;  Vincent et al., 2010; Coates and Ng, 2012).  Recently, learning such unsupervised representations has allowed in designing an effective clinical decision making system (Miotto et al., 2016). However, its not exactly clear on how to design a learning system for a continual and online learning setting using representations obtained through unsupervised learning, because we assume that we do not have access to data prior to the beginning of a learning task.

\section{Algorithm}

We consider a single-hidden layer neural network with a single output unit for presenting our algorithm. The parameters $ U \in \mathbb{R}^{m \times n} $ and $ W \in \mathbb{R}^{n} $ are the incoming and outgoing weights of the neural network where $ m $ is the number of input units and $ n $ is the number of hidden units. Each element of $ U $ is denoted as $ u_{ij} $ where $ i $ refers to the corresponding input unit and $ j $ refers to the hidden unit. Likewise, each element of $ W $ is denoted as $ w_{j} $. 

\begin{algorithm}
\algnewcommand\algorithmicinput{\textbf{INPUT:}}
\algnewcommand\INPUT{\item[\algorithmicinput]}
\newcommand\NoDo{\renewcommand\algorithmicdo{}}
\newcommand\ReDo{\renewcommand\algorithmicdo{\textbf{do}}}
\newcommand\NoThen{\renewcommand\algorithmicthen{}}
\newcommand\ReThen{\renewcommand\algorithmicthen{\textbf{then}}}

\caption{\text{Crossprop algorithm}}
\label{alg:crossprop}
\begin{algorithmic}[1]
\INPUT $ \alpha, \eta, m, n $ 
\State Initialize $ h_{ij} $ to 0
\State Initialize $ u_{ij} $ and $ w_{ij} $ as desired where $ i = 1, 2, \cdots, m; j = 1, 2, \cdots, n $
\For{each new example ($ X_{t}, y_{t}^{*} $) }
\State $ y \gets \sum_{j=1}^{n} \phi_{j, t}w_{j, t} $
\State $ \delta_{t} \gets y_{t}^{*} - y_{t} $
\For{$ j = 1, 2, \cdots, n $}
\For{$ i = 1, 2, \cdots, m $}
\State $ u_{ij, t + 1}  \gets u_{ij, t} + \alpha \delta_{t} \Big[ (1 - \eta)\phi_{j, t}h_{ij, t} + \eta w_{j, t}\frac{\partial \phi_{j, t}}{u_{ij, t}} \Big] $
\State $ h_{ij, t + 1} \gets h_{ij, t} \Big( 1 - \alpha (1 - \eta)\phi_{i, t}^{2} \Big) + \alpha\Big( \delta_{t} - \eta w_{j, t}\phi_{j, t} \Big) \frac{\partial \phi_{j, t}}{\partial u_{ij, t}} $
\EndFor
\State $ w_{j, t + 1} \gets w_{j, t} + \alpha \delta_{t} \phi_{j, t} $  
\EndFor
\EndFor
\end{algorithmic}
\end{algorithm}

Our proposed method is summarized as a pseudo-code in algorithm \ref{alg:crossprop} (and the code is available on github\footnote{https://github.com/ShangtongZhang/Crossprop}). A learning system (for simplicity, consider a single-hidden layer network), at time step $ t $, receives an example $ X_{t} \in \mathbb{R}^{m} $ where each element of this vector is denoted as $ x_{i, t} $.  This is mapped onto the hidden units through the incoming weight matrix $ U $ and a nonlinearity, like $ tanh $, $ sigmoid $ or $ relu $, is applied over this summed-product. The activations for each hidden unit for a given example at time step $ t $ using a $ \tanh $ activation function is expressed mathematically as, $ \phi_{j, t} = \tanh \Big( \sum_{i = 1}^{m} x_{i, t} u_{ij, t} \Big) $. These hidden units are successively mapped to form a scalar output $ y_{t} \in \mathbb{R} $ using the weights $ W $, which can be expressed as $ y_{t} = \sum_{j = 1}^{n} \phi_{j, t} w_{j, t} $.

Let $ \delta_{t}^{2} = \big( y_{t}^{*} - y_{t} \big)^{2} $ be a noisy objective function where $ y_{t}^{*} $ is the scalar target and $ y_{t} $ is the estimate made by an algorithm for an example at time step $ t $. The incoming and outgoing weights ($ U $ and $ W $) are incrementally learned after processing an example one after the other. 

The outgoing weights $ W $ are updated using the least mean squares (LMS) learning rule after processing an example at time step $ t $ as follows:
\begin{align*} 
    w_{j, t + 1} &= w_{j, t} - \frac{1}{2} \alpha \frac{\partial \delta^{2}_{t}}{\partial w_{j, t}} \\
    &= w_{j, t} - \alpha \delta_{t} \frac{\partial \delta_{t}}{\partial w_{j, t}} \\
    &= w_{j, t} - \alpha \delta_{t} \frac{\partial [ y^{*}_{t} - y_{t} ]}{\partial w_{j, t}} \\
    &= w_{j, t} + \alpha \delta_{t} \frac{\partial y_{t}}{\partial w_{j, t}} \\
    w_{j, t + 1} &= w_{j, t} + \alpha \delta_{t} \frac{\partial }{\partial w_{j, t}} \Big[ \sum_{i=1}^{n} \phi_{i, t}w_{i, t} \Big] \\
    w_{j, t + 1} &= w_{j, t} + \alpha \delta_{t} \phi_{j, t} \numberthis
    \label{eqn:w_update_1}
\end{align*}


We diverge from the conventional way (i.e., through {\em backprop}) for learning the incoming weights $ U $. Specifically, for learning the weights $ U $, we consider the influence of all the past values of $ U_{1}, U_{2}, \cdots U_{t} $ on the current error $ \delta_{t}^{2} $. We would like to learn the values of $ u_{ij, t + 1} $ by making an update using the partial derivative term $ \frac{\partial \delta_{t}^{2}}{\partial u_{ij}} $ where $ u_{ij} $ refers to all its past values.  

This is interesting because most of the current research on representation learning usually consider only the influence of the weight at the current time step $ u_{ij, t} $ on the squared error $ \delta_{t}^{2} $: $ \frac{\partial \delta_{t}^{2}}{\partial u_{ij, t}} $. This ignores the effects of the previous possible values of these weights on the squared error at the current time step. 

We now derive the update rule for the incoming weights as follows:
\begin{align*} 
   u_{ij, t + 1} &= u_{ij, t} - \frac{1}{2} \alpha \frac{\partial \delta_{t}^{2}}{\partial u_{ij}} \\
   &= u_{ij, t} -  \alpha \delta_{t} \frac{\partial [ y_{t}^{*} - y_{t} ]}{\partial u_{ij}} \\
   u_{ij, t+1} &= u_{ij, t} + \alpha \delta_{t} \frac{\partial y_{t}}{\partial u_{ij}} \numberthis \label{eqn:u_ij_update_1}
\end{align*}
Adapting the meta-gradient descent approach, that was introduced by Sutton (1992) and Schraudolph (1999), we derive the update rule for the incoming weights $ U $ as follows:
\begin{align*} 
   \frac{\partial y_{t}}{\partial u_{ij}} &=  \sum_{k} \frac{\partial y_{t}}{\partial w_{k, t}}\frac{\partial w_{k, t}}{\partial u_{ij}}  + \sum_{k} \frac{\partial y_{t}}{\partial \phi_{k, t}}\frac{\partial \phi_{k, t}}{\partial u_{ij}} \\
   &= \sum_{k} \frac{\partial y_{t}}{\partial w_{k, t}}\frac{\partial w_{k, t}}{\partial u_{ij}}  + \sum_{k} \frac{\partial y_{t}}{\partial \phi_{k, t}}\frac{\partial \phi_{k, t}}{\partial u_{ij, t}} \\
   \frac{\partial y_{t}}{\partial u_{ij}} &\approx \frac{\partial y_{t}}{\partial w_{j, t}}\frac{\partial w_{j, t}}{\partial u_{ij}}  + \frac{\partial y_{t}}{\partial \phi_{j, t}}\frac{\partial \phi_{j, t}}{\partial u_{ij, t}} \numberthis \label{eqn:chain_rule_1}
\end{align*}

Any error made during estimation of $ y_{t} $ by the learning system is attributed to both the outgoing weights of the features and to the activations of the hidden units. The approximations of $ \sum_{k} \frac{\partial y_{t}}{\partial w_{k, t}}\frac{\partial w_{k, t}}{\partial u_{ij}} \approx \frac{\partial y_{t}}{\partial w_{j, t}}\frac{\partial w_{j, t}}{\partial u_{ij}} $ and $ \sum_{k} \frac{\partial y_{t}}{\partial \phi_{k, t}}\frac{\partial \phi_{k, t}}{\partial u_{ij, t}} \approx \frac{\partial y_{t}}{\partial \phi_{j, t}}\frac{\partial \phi_{j, t}}{\partial u_{ij, t}} $ are reasonable because the primary effect on the input weight $ u_{ij} $ will be through the corresponding output weight $ w_{j, t} $ and feature $ \phi_{j, t} $. 


By defining $ h_{ij, t} = \frac{\partial w_{j, t}}{\partial u_{ij}} $, we can obtain a simple form for eqn. (\ref{eqn:chain_rule_1}):
\begin{align*} 
    \frac{\partial y_{t}}{\partial u_{ij}} &\approx \frac{\partial y_{t}}{\partial w_{j, t}}\frac{\partial w_{j, t}}{\partial u_{ij}} + \frac{\partial y_{t}}{\partial \phi_{j, t}}\frac{\partial \phi_{j, t}}{\partial u_{ij, t}}  \\
    &= \Big( \frac{\partial}{\partial w_{j, t}} \sum_{k} \phi_{k, t} w_{k, t} \Big)h_{ij, t} + \frac{\partial y_{t}}{\partial \phi_{j, t}}\frac{\partial \phi_{j, t}}{\partial u_{ij, t}} \\
    \frac{\partial y_{t}}{\partial u_{ij}} &\approx \phi_{j, t} h_{ij, t} + \frac{\partial y_{t}}{\partial \phi_{j, t}}\frac{\partial \phi_{j, t}}{\partial u_{ij, t}} \numberthis \label{eqn:chain_rule_2}
\end{align*}

The partial derivative $ \frac{\partial y_{t}}{\partial \phi_{j, t}}\frac{\partial \phi_{j, t}}{\partial u_{ij, t}} $ is the conventional backprop update. However, in our proposed algorithm, we have an additional update term $ \phi_{j, t} h_{ij, t} $ that captures the dependencies of all the previous values of $ u_{ij} $ on the current estimate $ y_{t} $ and on the current squared error $ \delta_{t}^{2} $.

\def\first#1#2{\frac{\partial #1}{\partial #2}}
$ h_{ij, t} $ is an additional memory parameter corresponding to the input weight $ u_{ij, t} $ and can be written as a recursive update equation as follows: 
\begin{align*} 
   h_{ij, t + 1} &\approx \first{w_{j, t + 1}}{u_{ij}} \\
   &= \first{}{u_{ij}} \Big[ w_{j, t} + \alpha \delta_{t} \phi_{j, t} \Big] \\
   &= \first{w_{j, t}}{u_{ij}} + \alpha\first{}{u_{ij}} \Big[ \delta_{t}\phi_{j, t} \Big]\\
   &= h_{ij, t} + \alpha \delta_{t} \first{\phi_{j, t}}{u_{ij, t}} + \alpha \first{\delta_{t}}{u_{ij}} \phi_{j, t} \\
   &\approx h_{ij, t} + \alpha \delta_{t} \first{\phi_{j, t}}{u_{ij, t}} + \alpha \first{\delta_{t}}{y_{t}}\first{y_{t}}{u_{ij, t}}\phi_{j, t} \\
   &\approx h_{ij, t} + \alpha \delta_{t} \first{\phi_{j, t}}{u_{ij, t}} + \alpha \first{\delta_{t}}{y_{t}} \frac{\partial y_{t}}{\partial w_{j, t}}\frac{\partial w_{j, t}}{\partial u_{ij}}\phi_{j, t} + \alpha \first{\delta_{t}}{y_{t}}\frac{\partial y_{t}}{\partial \phi_{j, t}}\frac{\partial \phi_{j, t}}{\partial u_{ij, t}}\phi_{j, t} \\
   &= h_{ij, t} + \alpha \delta_{t} \first{\phi_{j, t}}{u_{ij, t}} - \alpha \frac{\partial y_{t}}{\partial w_{j, t}}\frac{\partial w_{j, t}}{\partial u_{ij}}\phi_{j, t} - \alpha \frac{\partial y_{t}}{\partial \phi_{j, t}}\frac{\partial \phi_{j, t}}{\partial u_{ij, t}}\phi_{j, t} \\
   &= h_{ij, t} + \alpha \delta_{t} \first{\phi_{j, t}}{u_{ij, t}} - \alpha \phi_{j, t}^{2} h_{ij, t} - \alpha w_{j, t}\phi_{j, t}\frac{\partial \phi_{j, t}}{\partial u_{ij, t}} \\
   h_{ij, t} &\approx h_{ij, t} \Big( 1 - \alpha \phi_{j, t}^{2} \Big) + \alpha \Big( \delta_{t} - w_{j, t}\phi_{j, t} \Big) \first{\phi_{j, t}}{u_{ij, t}} \numberthis
\end{align*}



By substituting eqn. (\ref{eqn:chain_rule_2}) in eqn. (\ref{eqn:u_ij_update_1}), we define a recursive update equation for the weights $ u_{ij, t} $ and thereby summarize the complete algorithm as follows:
\begin{align}\label{eqn:alg_updates}
\begin{split}
    u_{ij, t + 1} &= u_{ij, t} + \alpha \delta_{t} \Big[ \phi_{j, t} h_{ij, t} + w_{j, t} \first{\phi_{j, t}}{u_{ij, t}} \Big] \\
    h_{ij, t + 1} &= h_{ij, t} \Big( 1 - \alpha \phi_{j, t}^{2} \Big) + \alpha \Big( \delta_{t} - w_{j, t}\phi_{j, t} \Big) \first{\phi_{j, t}}{u_{ij, t}} \\
    w_{i, t + 1} &= w_{i, t} + \alpha \delta_{t} \phi_{i, t}
\end{split}
\end{align}

Depending on the nonlinearity used for the hidden units, $ \frac{\partial \phi_{j, t}}{\partial u_{ij, t}} $ can be reduced to a closed-form equation.

For instance, if a logistic function is used, then $ \phi_{j, t} = \sigma \Big( \sum_{i=1}^{m}x_{i, t}u_{ij, t}  \Big) $,
\begin{align*} 
   \frac{\partial \phi_{j, t}}{\partial u_{ij, t}} &= \frac{\partial \phi_{j, t}}{\partial u_{ij, t}} \\
   &= \frac{\partial}{\partial u_{ij, t}} \sigma \Big( \sum_{i=1}^{m} x_{i, t}u_{ij, t} \Big) \\ 
   \frac{\partial \phi_{j, t}}{\partial u_{ij, t}} &= \phi_{j, t} \Big( 1 -  \phi_{j, t} \Big) x_{i, t}
\end{align*}

Another frequently used activation function is $ \tanh $, which implies that $ \phi_{j, t} = \tanh \Big( \sum_{i=1}^{m} x_{i, t}u_{ij, t} \Big) $,
\begin{align*} 
   \frac{\partial \phi_{j, t}}{\partial u_{ij, t}} &= \frac{\partial \phi_{j, t}}{\partial u_{ij, t}} \\
   &= \frac{\partial}{\partial u_{ij, t}} \tanh \Big( \sum_{i=1}^{m} x_{i, t}u_{ij, t} \Big) \\ 
   \frac{\partial \phi_{j, t}}{\partial u_{ij, t}} &= \Big( 1 -  \phi_{j, t}^2 \Big) x_{i, t}
\end{align*}

We could also introduce a weighting factor $ \eta \in [0, 1] $ in eqn. (\ref{eqn:chain_rule_2}), which allows in smoothly mixing backprop and meta-gradient updates,
\begin{align*} 
    \frac{\partial y_{t}}{\partial u_{ij}} \approx (1 - \eta)\phi_{j, t} h_{ij, t} + \eta \frac{\partial y_{t}}{\partial \phi_{j, t}}\frac{\partial \phi_{j, t}}{\partial u_{ij, t}}
\end{align*} 
which results in the following update equations for learning the weights $ U $ and $ W $ of the neural network:  
\begin{align}\label{eqn:alg_updates_eta}
\begin{split}
    u_{ij, t + 1} &= u_{ij, t} + \alpha \delta_{t} \Big[ (1 - \eta) \phi_{j, t} h_{ij, t} + \eta w_{j, t} \first{\phi_{j, t}}{u_{ij, t}} \Big] \\
    h_{ij, t + 1} &= h_{ij, t} \Big( 1 - \alpha (1 - \eta) \phi_{j, t}^{2} \Big) + \alpha \Big( \delta_{t} - \eta w_{j, t}\phi_{j, t} \Big) \first{\phi_{j, t}}{u_{ij, t}} \\
    w_{i, t + 1} &= w_{i, t} + \alpha \delta_{t} \phi_{i, t}
\end{split}
\end{align}

The algorithm that was derived and presented in eqns. (\ref{eqn:alg_updates_eta}) and (\ref{eqn:alg_updates}) are computationally expensive when there are more number of outgoing weights per hidden unit (here, this means that there are more than one output unit). Specifically, when there are $ k $ output units, then $ \delta_{t} $ becomes a $ k $-dimensional vector with dimensions equal to that of the output units. This leads to a large computational cost involved in computing $ h_{ij, t} $, which can be avoided by approximating the $ h_{ij, t} $ parameter. The approximation involves in accumulating the error assigned to each of the hidden units through its outgoing weights and using this to compute the update term. This approximated algorithm is referred to as crossprop-approx. in our experiments and has the following update equations:

\begin{align}\label{eqn:alg_updates_eta_1}
\begin{split}
    u_{ij, t + 1} &= u_{ij, t} + \alpha \sum_{k} \delta_{k, t} \Big[ (1 - \eta) \phi_{j, t} h_{jk, t} + \eta w_{jk, t} \Big]\first{\phi_{j, t}}{u_{ij, t}} \\
    h_{jk, t + 1} &= h_{jk, t} \Big( 1 - \alpha (1 - \eta) \phi_{j, t}^{2} \Big) + \alpha \Big( \delta_{k, t} - \eta w_{jk, t} \phi_{j, t} \Big) \\
    w_{jk, t + 1} &= w_{jk, t} + \alpha \delta_{k, t} \phi_{j, t}
\end{split}
\end{align}

\section{Experiments and Results}
Here we empirically investigate whether crossprop is effective in finding useful representations for continual learning tasks and compare them with backprop and its many (such as adding momentum, RMSProp and ADAM). By continual learning tasks, we refer to an experiment setting where supervised training examples are generated and presented to a learning system from a sequence of related tasks. Moreover, the learning system does not know when the task is switched. 

\subsection{GEOFF Tasks}
The GEneric Online Feature Finding (GEOFF) problem was first introduced by Sutton (2014) as a generic, synthetic, feature-finding test bed for evaluating different representation learning algorithms. The primary advantage of this test bed is that infinitely many supervised-learning tasks can be generated without any experimenter bias. 

The test bed consists of a single hidden layer neural network, called the {\em target network}, with a real-valued scalar output. Each input example, $ X_{t} \in \{ 0, 1 \}^{m} $, is a $ m $-dimensional binary input vector where each element in the vector can take a value of 0 or 1. The hidden layer consists of $ n $ Linear Threshold Units (LTUs), $ \boldsymbol{\phi}_{t}^{*} \in \{ 0, 1 \}^{n} $, with a threshold parameter of $ \beta $. The $ \beta $ parameter controls the sparsity in the hidden layer. The weights $ U^{*} \in \{ -1, +1 \}^{m \times n} $ maps the input vector to the hidden units and the weights $ W^{*} \in \{ -1, 0, +1 \}^{n} $ linearly combine the LTUs (features) to produce a scalar target output $ y^{*} $. The weights $ U^{*} $ and $ W^{*} $ are generated using a uniform probability distribution and remain fixed throughout a task, representing a stationary function mapping a given input vector $ X_{t} $ to a scalar target output $ y_{t}^{*} $. The input vector $ X_{t} $ is generated randomly using a uniform probability distribution. For each input vector, this target network is used to produce a scalar target output $ y_{t}^{*} = \sum_{i=1}^{n} \phi_{i, t}^{*} w^{*}_{i} + \mathcal{N}(0, 1) $. For our experiments, we fix $ m = 20 $, $ n = 1000 $ and $ \beta = 0.6 $ (the parameters of the target network).

\textbf{Experiment setup.} For our experiments, we create an instance of the GEOFF task. This is called Task A and use this to generate a set of 5000 examples. These examples are then used for training the learning systems in an online manner, where each example is processed once and then discarded. After processing the examples from Task A, we generate a Task B by randomly choosing and regenerating 50\% of the outgoing target weights $ W^{*} $. A set of 5000 training examples are generated for training from this modified task. Similarly, after processing the examples from Task B, Task C is produced which is used for generating another 5000 training examples. The learning systems learn online from training examples produced by a sequence of related tasks (Tasks A, B \& C) where the representations learned from one can be used for solving the other tasks. It is important to point out here that all these tasks share the same feature representation (i.e. the weights $ U ^ {*} $ remain fixed throughout) and the learning system can leverage from its previous learning experiences. 

This experiment was setup from a continual learning perspective where a learning system will experience examples generated from a sequence of related tasks and the learning from one task can help in learning other similar tasks. The step-size for all the algorithms was fixed at a constant value ($ \alpha = 0.0005 $) as the objective here is to show how the features are learned by different algorithms for a sequence of related learning tasks. The learning system consisted of a single hidden layer neural network with a single output unit. It had 20 input units and 500 hidden units using $ \tanh $ activation function. The squared error function was used for learning the parameters of this network. These were the parameters of the learning network used for evaluating multiple algorithms.

\begin{figure*}[ht!]
\begin{adjustwidth}{-0.5cm}{}
\centering
\begin{tabular}{cc} 
\subfloat[]{\includegraphics[keepaspectratio,width=6cm,height=7cm]{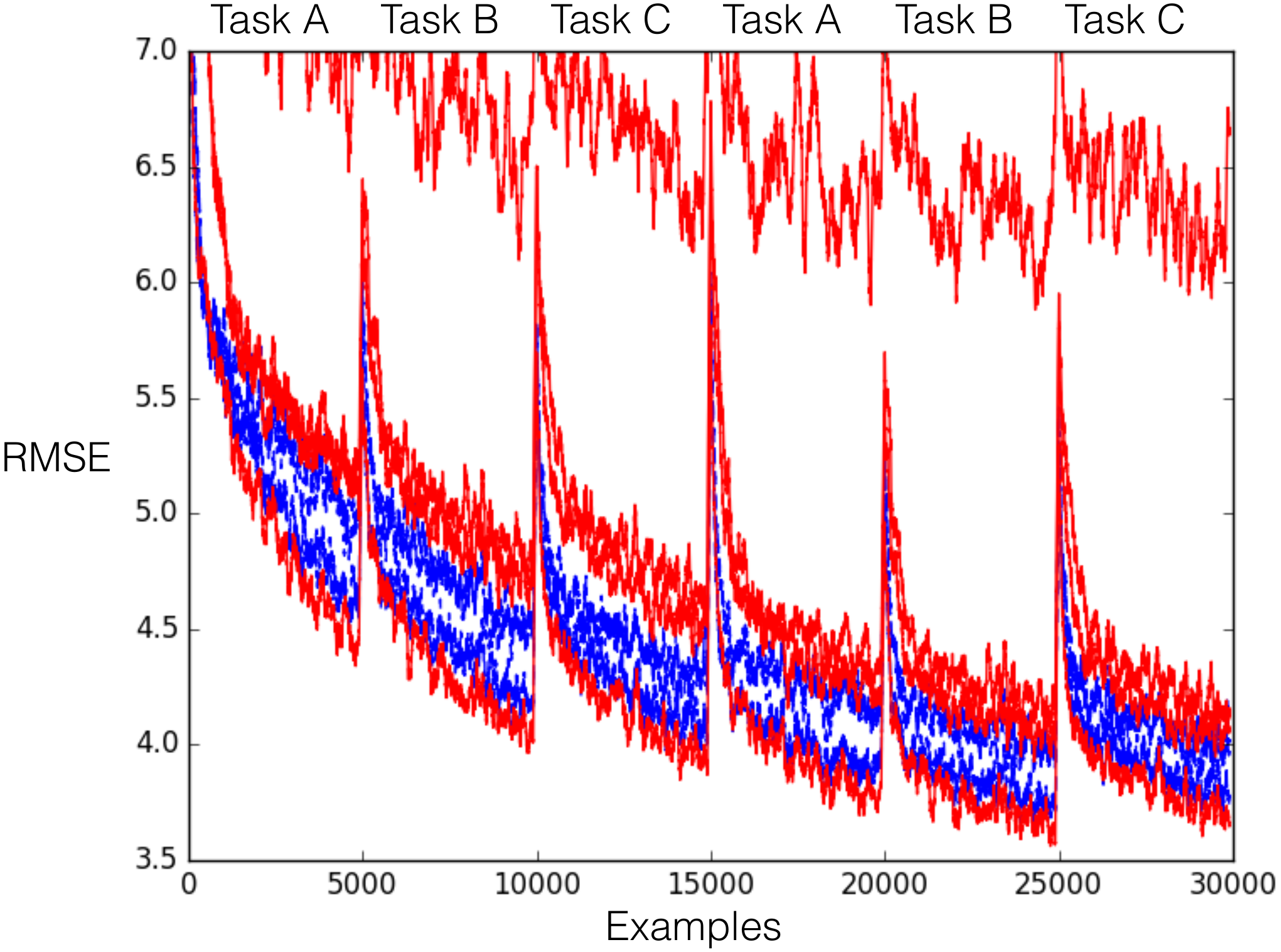}} & 
\subfloat[]{\includegraphics[keepaspectratio,width=6.4cm,height=7cm]{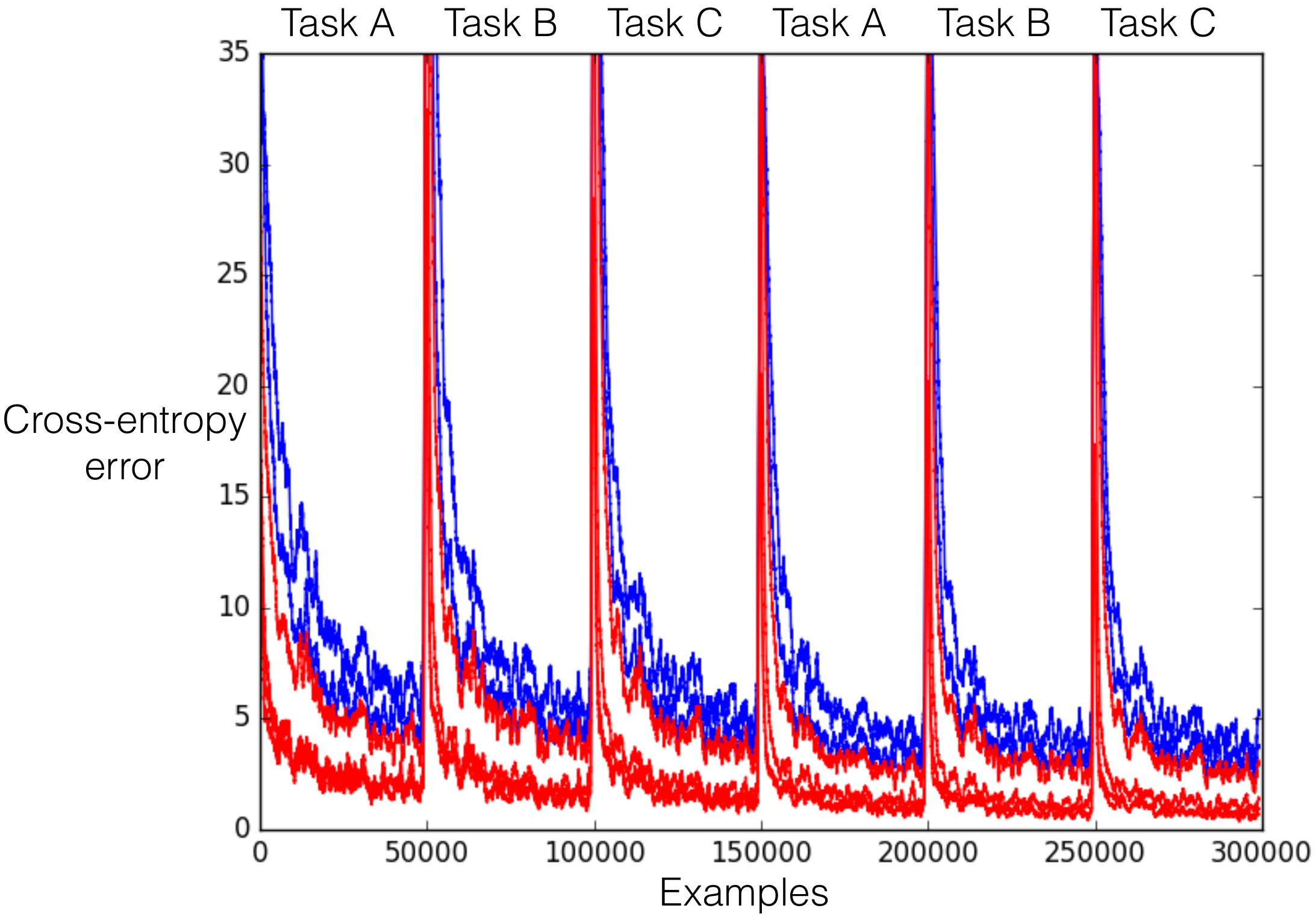}}
\end{tabular}
\end{adjustwidth}
\caption{The learning curves of crossprop (i.e., crossprop, crossprop-approx with $ \eta = 0, 0.5 $) are colored blue and backprop (backprop, momentum, RMSProp, ADAM) are colored red. The learning systems do not know that the task has changed. Figure (a) shows the learning curves on a series of GEOFF tasks and figure (b) shows the learning curves on a series of MNIST tasks. The learning curves on figure (a) are averaged over 30 independent runs where each run used different target networks from generating the training examples. The learning curves on figure (b) are from a single run where the training set of MNIST was used. Also, on the MNIST tasks, only crossprop-approx. was evaluated.}
 \label{fig:learning_curves}
\end{figure*}

\begin{figure*}[ht!]
\begin{adjustwidth}{-0.5cm}{}
\centering
\begin{tabular}{cc} 
\subfloat[]{\includegraphics[keepaspectratio,width=6cm,height=7cm]{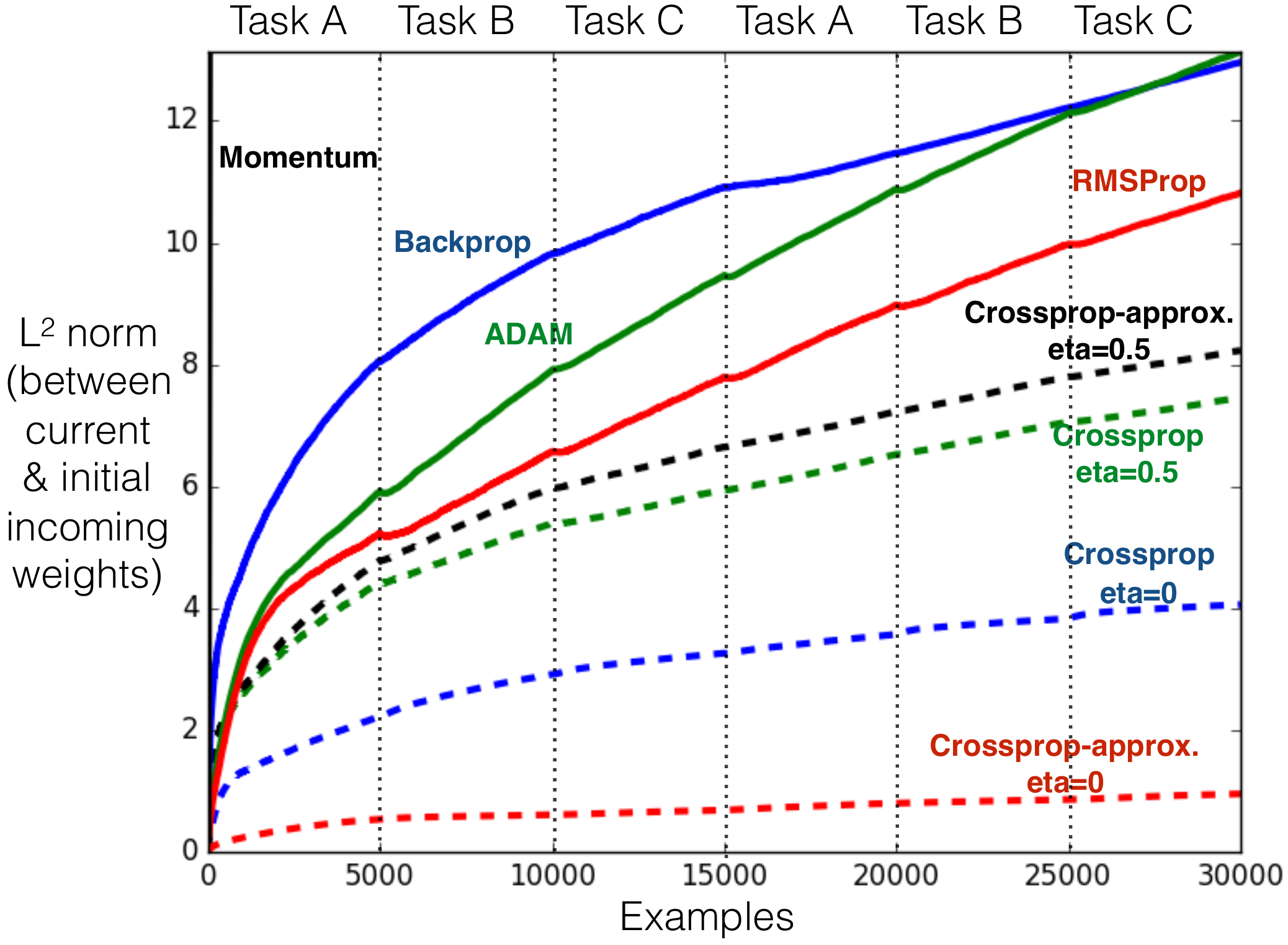}} & 
\subfloat[]{\includegraphics[keepaspectratio,width=6.1cm,height=7cm]{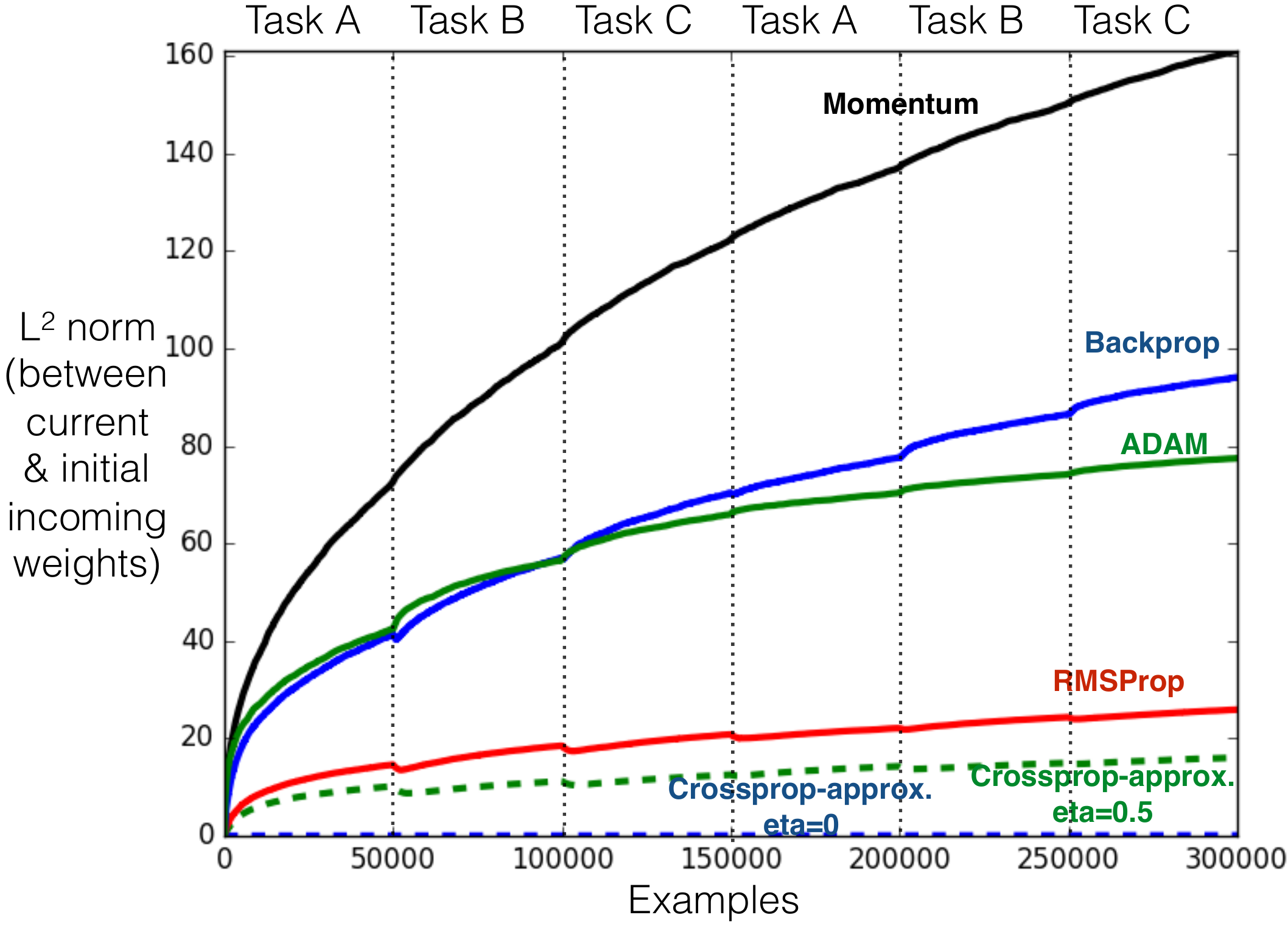}}
\end{tabular}
\end{adjustwidth}
\caption{The plot shows the change in the incoming weights $ U $ after processing each example. Specifically, the plot shows {\em l\textsuperscript{2}} norm between the incoming weights $ U $ after processing the n\textsuperscript{th} example and its initialized value for different learning algorithms. From the plot, it can be observed that crossprop tends to change the incoming weights the least even when the task is significantly changed, implying that crossprop tends to find a reusable feature representation that can sufficiently solve the sequence of tasks that it experiences. On the other hand, backprop tends to significantly {\em relearn} the feature representation throughout the experiment even when the task can be solved by leveraging from previous learning experiences.}
 \label{fig:U_weights}
\end{figure*}

\begin{figure*}[ht!]
\begin{adjustwidth}{-0.5cm}{}
\centering
\begin{tabular}{cc} 
\subfloat[]{\includegraphics[keepaspectratio,width=6cm,height=7cm]{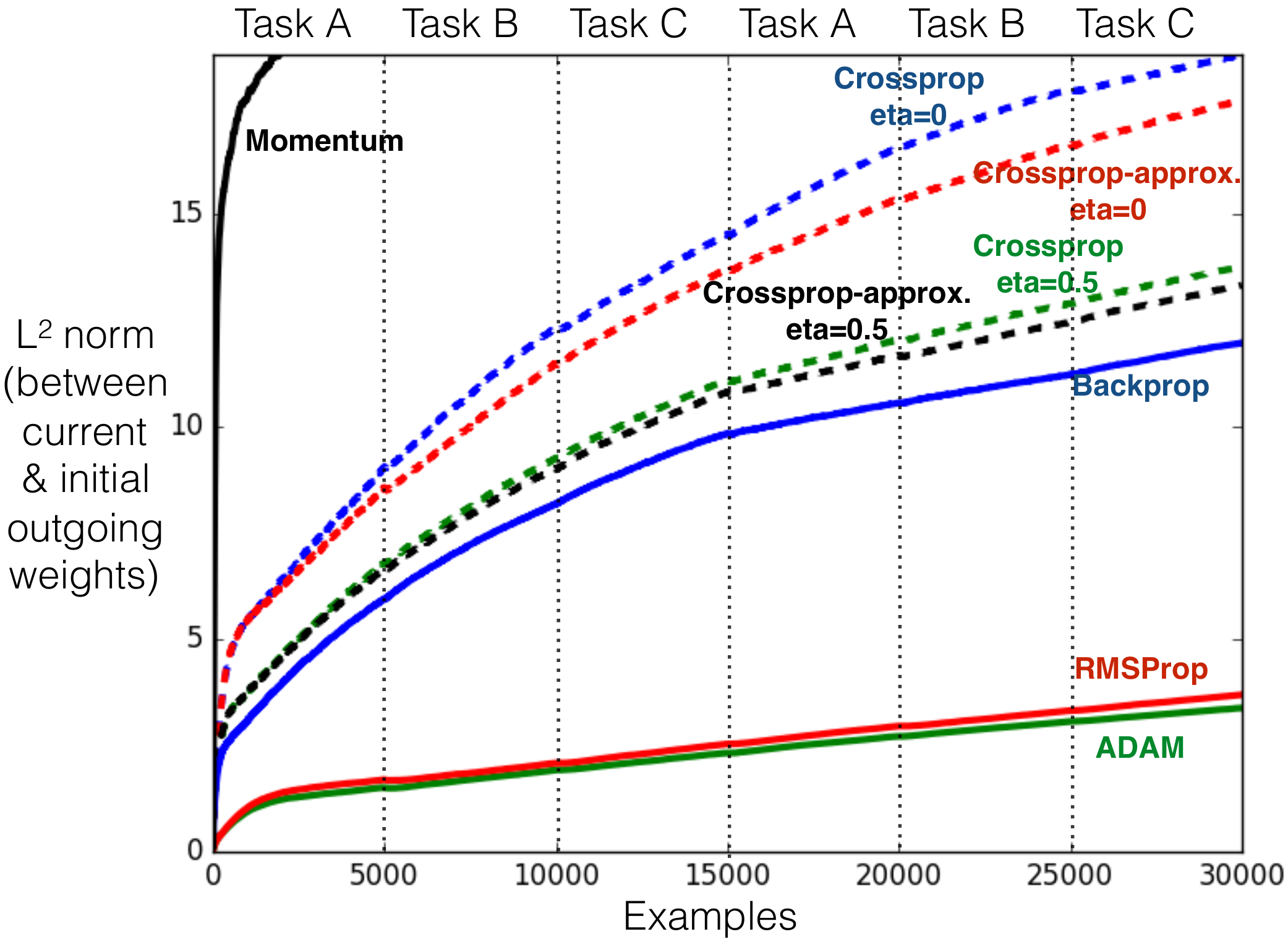}} & 
\subfloat[]{\includegraphics[keepaspectratio,width=6.05cm,height=7cm]{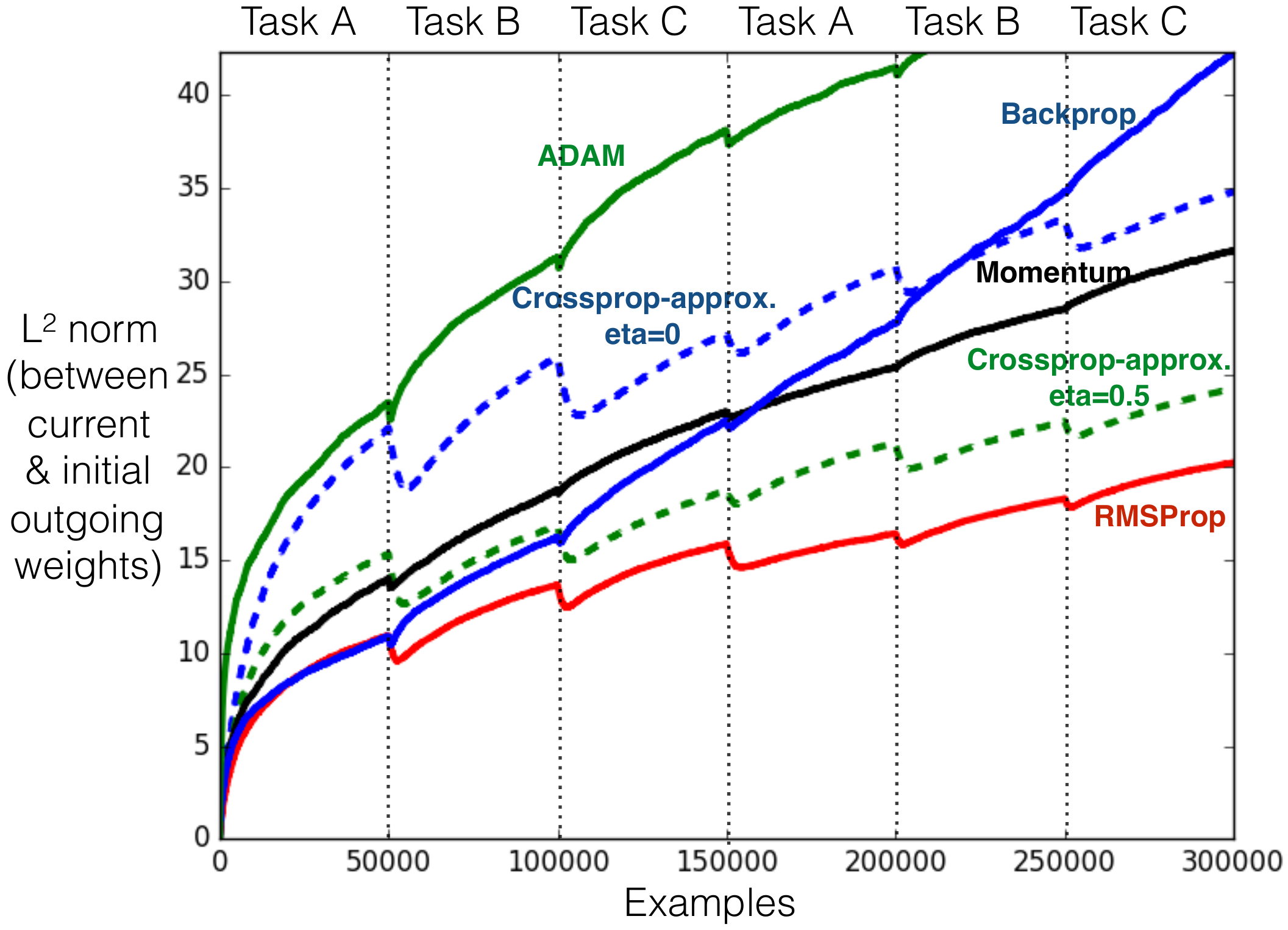}}
\end{tabular}
\end{adjustwidth}
\caption{The plot shows the change in the outgoing weights $ W $ after processing each example. It shows the {\em l\textsuperscript{2}} norm between the outgoing weights $ W $ after processing the n\textsuperscript{th} example and its initialized value for different learning algorithms. From the plot, it can be observed that crossprop tends to change the outgoing weights the most as it is needed to map the feature representation to an estimate $ y_{t} $. In backprop, each parameter independently minimizes the error and because of this, each parameter race with each other in reducing the error without coordinating their efforts. So, it tends to change its feature representation for accommodating a new example.}
 \label{fig:W_weights}
\end{figure*}

\textbf{Results.} We compare the behavior of crossprop with backprop and its variations on the sequence of related tasks generated using the GEOFF testbed. Figure \ref{fig:learning_curves} (a) shows the learning curve for different algorithms. After every 5000 examples, the task switches to a new and related task as previously described. It is important to note here that the learning system does not know that the task has changed. 

The learning curves show that crossprop reaches a similar asymptotic value to that of backprop, implying that the introduced algorithm produces a similar solution as backprop. In terms of asymptotic values, backprop achieves a significantly better asymptotic value compared to crossprop and the other variations of backprop. However, it is interesting to note that these learning algorithms approach the solution differently. 

Figure \ref{fig:U_weights} (a) shows the euclidean norm ($ l^{2} $ norm) between the weights $ U $ after processing the n\textsuperscript{th} training example and the initialized value of the same weights. Though all the algorithms reach similar asymptotic values, the way backprop achieves this is clearly different from that of crossprop. Backprop tends to frequently modify the features even though it has seen examples that are generated using a previously learned function. Specifically, backprop fails to leverage from its previous learning experiences in solving new tasks even when it is possible. Because of this backprop tends to take a lot of time in finding a feature representation which can sufficiently solve this continual problem. This is clearly not the case with crossprop. Our proposed algorithm tends to find a feature representation much quicker than backprop that can sufficiently solve the sequence of continual problems and reuses this for solving new tasks that it encounters in the future. 

Figure \ref{fig:W_weights} (a) shows the euclidean norm between the weights $ W $ after processing the n\textsuperscript{th} example and the initialized value of the same weights. Because crossprop tends to find the set of features much quicker than backprop and reuses these features while solving a new task, it reduces the error by moving the outgoing weights rather than modifying its feature representation. Furthermore, all the tasks presented to the learning system can be solved by using a single feature representation and from our plots, it can be clearly seen that crossprop recognizes this.

\subsection{MNIST Tasks}

The MNIST dataset of handwritten digits was introduced by LeCun, Cortes and Burges (1998). Though the MNIST dataset is old, it is still viewed as a standard supervised learning benchmark task for testing out new learning algorithms (Sironi et al., 2015; Papernot et al., 2016). 

The dataset consists of grayscale images each with $ 28 \times 28 $ dimensions. These images are obtained from handwritten digits and their corresponding labels denote the supervised learning target for a given image. The objective of a learning system in a MNIST task would be to learn a mapping function that maps each of these images to a label.

\textbf{Experiment setup.} We adapt the MNIST dataset to a continual learning setting, where in each task the label for the training images is shifted by one. For example, Task A uses the standard MNIST training images and their labels, Task B uses the same training examples as Task A, but now the labels get shifted by one. Similarly, for Task C the label for the training examples get further shifted by one. As in our previous experiment, we fix the step-size ($ \alpha = 0.0005 $) for the different algorithms as our objective here is to study how the representations are learned between these algorithms for a continual learning setting, where the learning system experiences examples from a sequence of related tasks. The learning system consisted of a single hidden layer neural network with 784 input units, 1024 hidden units and 10 output units. The hidden units used a $ \tanh $ activation function and the output units used a softmax activation function. The cross-entropy error function was used for training the network. 

\textbf{Results.} Figure \ref{fig:learning_curves} (b) shows the learning curves for all the methods evaluated on the MNIST tasks. As observed in the GEOFF tasks, the learning curves for the different algorithms converge to almost similar points which means that all the methods reach similar solutions. However, ADAM and RMSProp achieves a significantly better asymptotic error value compared to the other learning algorithms.

Figures \ref{fig:U_weights} (b) and \ref{fig:W_weights} (b) show the euclidean norm of the change in weights $ U $ and $ W $ respectively. As seen in our previous experiments, crossprop tends to find the features much quicker than backprop and its variations. Also, crossprop tends to reuse these features in solving the new tasks that it faces. It is interesting to observe that backprop does not seem to settle down on a good feature representation for solving a sequence of continual learning problems. It tends to na{\"i}vely {\em unlearn} and {\em relearn} its feature representation even when the tasks are similar to each other and can be solved by using the feature representation learned from the first task. Specifically, backprop does not seem to leverage its previous learning experiences while encountering a new task. 

\begin{figure*}[ht!]
\begin{adjustwidth}{-2cm}{}
\centering
\begin{tabular}{cc} 
\subfloat[]{\includegraphics[keepaspectratio,width=8cm,height=8cm]{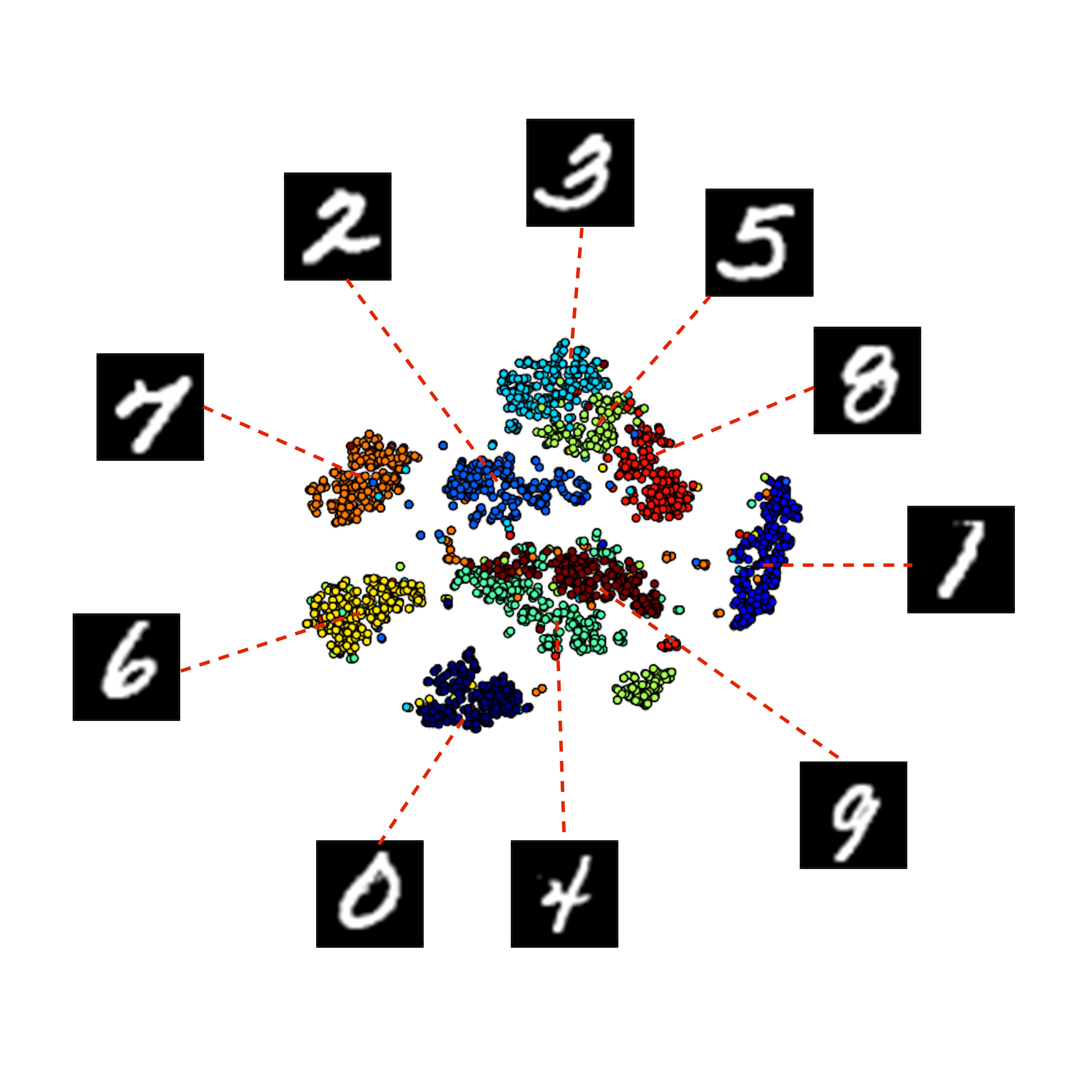}} & 
\subfloat[]{\includegraphics[keepaspectratio,width=8cm,height=8cm]{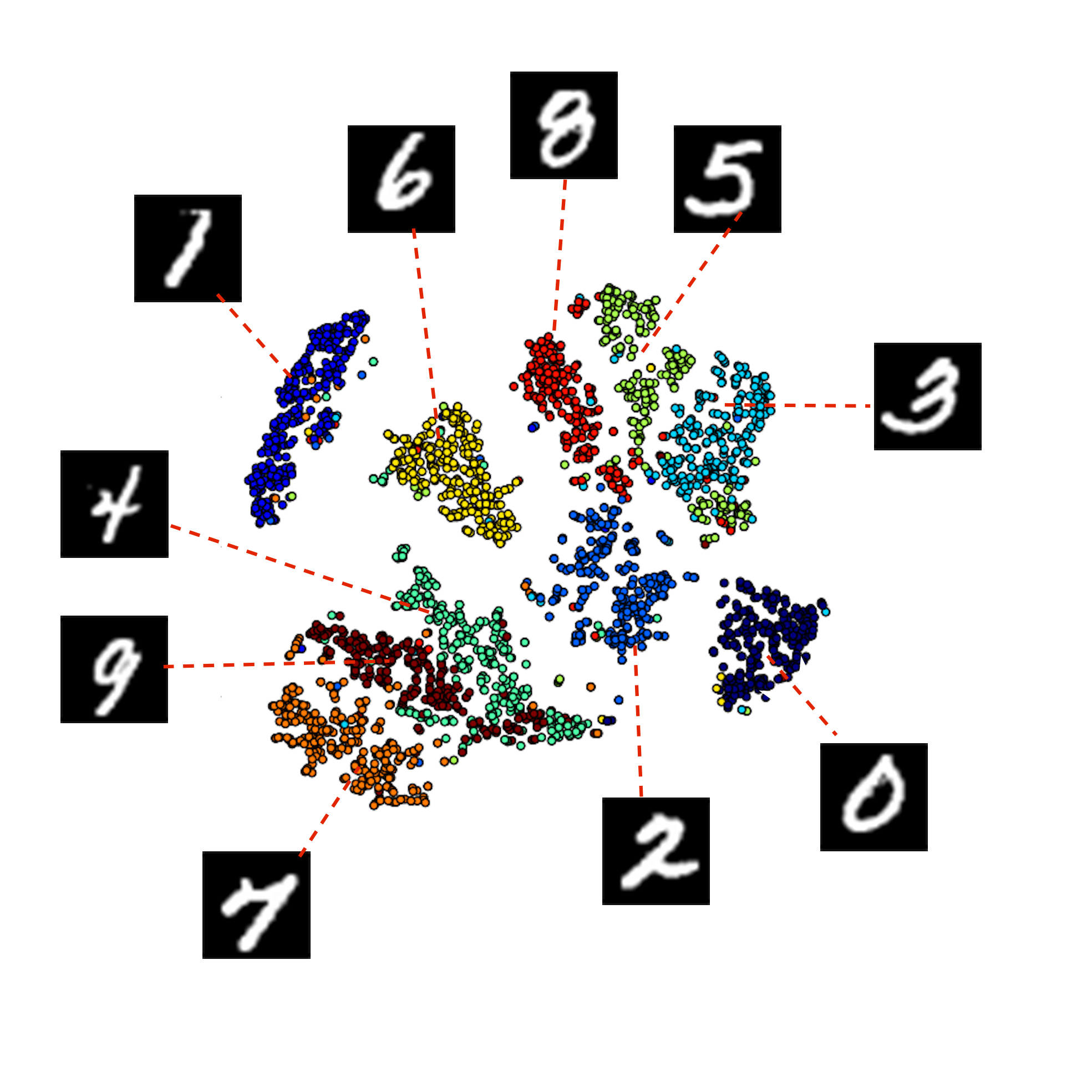}}
\end{tabular}
\end{adjustwidth}
\caption{Backprop and crossprop seem to learn similar feature representations, by clustering together similar examples. The plot shows the visualizations of the features (i.e. activations of the hidden units) learned by backprop (a) and crossprop (b) on the standard MNIST task. Both the learning algorithms were trained online on the MNIST dataset and the parameters learned by these algorithms were used for generating these visualizations. The $ \eta $ was set to 0 for crossprop in order to draw out differences between the conventional and the meta-gradient descent approach for learning the features. The plot was generated by using 2500 training examples, uniformly sampled from the MNIST dataset.}
 \label{fig:tsne}
\end{figure*}

\section{Visualizing the learned features}

We visualize the features that are obtained while training the learning systems using crossprop and backprop. These visualization are obtained using the t-SNE approach, which was developed by Maaten and Hinton (2008) for visualizing high-dimensional data by giving each datapoint a location in a two or three-dimensional map. Here, we show only the two-dimensional map generated using the features learned by the different learning algorithms.

The features learned by backprop and crossprop (with $ \eta $ set to 0) on a standard MNIST task are plotted in Figures \ref{fig:tsne} (a) and (b). From the visualizations, it can be observed that both these algorithms produce similar feature representations on the task. Both these algorithms learn a feature representation that clusters examples according to their labels. There does not seem to be much of a difference between them by looking at their features. 

\section{Discussions}

Neural networks and backprop form a powerful, hierarchical feature learning paradigm. Designing deep neural network architectures has allowed many learning systems to achieve levels of performance comparable to that of humans in many domains. Many of the recent research works, however, fail to notice or ignore the fundamental issues that are present with backprop, even though it is important to address them. 

Some research works even tend to provide ad-hoc solutions to overcome these fundamental problems posed by backprop, but these are usually not scalable to general domains. Over time, many modifications were introduced to backprop, but they still fail to address the fundamental issue with backprop, which is that backprop tends to interfere with its previously learned representations in order to accommodate a new example. This prevents in directly applying backprop to continual learning domains, which is critical for achieving Artificial Intelligence (Ring, 1997; Kirkpatrick et al. 2017).

In a continual learning setting, a learning system needs to progressively learn and hierarchically accumulate knowledge from its experiences, using them to solve many difficult, unseen tasks. In such a setting, it is not desirable to have a learning system that na{\" i}vely {\em unlearns} and {\em relearns} even when it sees a task that can be solved by reusing its learning from its past experiences. Particularly, for a continual learning setting, it is necessary to have a learning system that can hierarchically build knowledge from its previous experiences and use them in solving a completely new and unseen task.

In this paper, we present two continual learning tasks that were adapted from standard supervised learning domains: the GEOFF testbed and MNIST dataset. On these tasks, we evaluate backprop and its variations (momentum, RMSProp and ADAM). We also evaluate our proposed meta-gradient descent approach for learning the features in a neural network, called crossprop. We show that backprop (and its variations) tends to relearn its feature representations for every task, even when these tasks can be solved by reusing the feature representation learned from previous experiences. Crossprop, on the other hand, tends to reuse its previously learned representations in tackling new and unseen tasks. The process of consistently failing to leverage from previous learning experiences is not particularly desirable in a continual learning setting which prevents in directly applying backprop to such settings. Addressing this particular issue is the primary motivation for our work. 

As an immediate future work, we would like to study the performances of this meta-gradient descent approach on deep neural networks and comprehensively evaluate them on more difficult benchmarks, like IMAGENET (Deng et al., 2009) and the Arcade Learning Environment (Bellemare et al., 2013).

\section{Conclusions}

In this paper, we introduced a meta-gradient descent approach, called crossprop, for learning the incoming weights of hidden units in a neural network and showed that such approaches are complementary to backprop, which is the popular algorithm for training neural networks. We also show that by using crossprop, a learning system can learn to reuse the learned features for solving new and unseen tasks. However, we see this as the first general work towards comprehensively addressing and overcoming the fundamental issues posed by backprop, particularly for continual learning domains.

\end{document}